\documentclass[fleqn,10pt]{wlscirep}
\usepackage[utf8]{inputenc}
\usepackage[T1]{fontenc}

\usepackage{tabularx}

\newcolumntype{C}{>{\centering\arraybackslash}X}

\title{Adaptive EEG‑based stroke diagnosis with a GRU–TCN classifier and deep Q‑learning thresholding}

\author[1,2,*,+]{Shakeel Abdulkareem}
\author[2,+]{Bora Yimenicioglu}
\author[2]{Khartik Uppalapati}
\author[1]{Aneesh Gudipati}
\author[3]{Adan Eftekhari}
\author[3]{Saleh Yassin}

\affil[1]{George Mason University, College of Science, Fairfax, VA, USA}
\affil[2]{Raregen Youth Network, Translational Medical Research Department, Oakton, VA, USA}
\affil[3]{Harvard University, Cambridge, MA, USA}

\affil[*]{Correspondence: \href{mailto:sabdulk2@gmu.edu}{sabdulk2@gmu.edu}}
\affil[+]{These authors contributed equally to this work.}


\begin{abstract}
Rapid triage of suspected stroke needs accurate, bedside-deployable tools; electroencephalography (EEG) is promising but remains underused for initial diagnosis. We developed an adaptive multitask EEG classifier that converts 32-channel signals to power spectral density features (Welch’s method), uses a recurrent–convolutional network (GRU–TCN) to predict stroke type (healthy, ischemic, hemorrhagic), hemispheric lateralization, and severity, and employs a deep Q-network (DQN) to adjust decision thresholds in real time. In a patient-wise split of the UCLH Stroke EIT/EEG dataset (44 recordings; $\sim$26 acute stroke, 10 controls), the primary outcome was stroke-type performance; secondary outcomes were severity and lateralization. The baseline GRU–TCN achieved 89.3\% accuracy (F1 92.8\%) for stroke type, $\sim$96.9\% (F1 95.9\%) for severity, and $\sim$96.7\% (F1 97.4\%) for lateralization. With DQN threshold adaptation, stroke-type accuracy increased to $\sim$98.0\% (F1 97.7\%). We also evaluated robustness on an independent, low-density EEG cohort (ZJU4H); comparative results and paired patient-level statistics are reported in the Results. Analyses followed STARD 2015 guidance for diagnostic accuracy studies (index test: GRU–TCN+DQN; reference standard: radiology/clinical diagnosis; patient-wise evaluation). Adaptive thresholding shifts the operating point to clinically preferred sensitivity–specificity trade-offs, while integrated scalp-map and spectral visualizations support interpretability.
\end{abstract}

\begin{document}

\flushbottom
\maketitle
%
%
\thispagestyle{empty}

\section*{Introduction}
Stroke remains a leading cause of death and long-term disability worldwide, and timely identification of stroke type and lesion laterality is critical for directing acute therapies \cite{feigin2021stroke,powers2019ais}. While computed tomography (CT) and magnetic resonance imaging (MRI) are the reference standards for etiologic classification, access, workflow, and transport constraints can delay decisions in prehospital and early in-hospital settings \cite{powers2019ais}. Surface electroencephalography (EEG) is portable, fast to acquire, and sensitive to acute neurophysiological disturbances after cerebrovascular injury, making it a plausible adjunct for early triage \cite{sutcliffe2022,wu2016}.

Existing EEG-based stroke classifiers typically rely on fixed feature sets and static decision thresholds. Such approaches can be brittle under patient heterogeneity, device differences, and variable signal quality, leading to operating points that do not generalize across clinical contexts \cite{wu2016}. Deep learning has improved EEG decoding and provided richer representations for clinical neurophysiology, but most models still apply a single, pre-specified threshold at inference \cite{schirrmeister2017hbm}. In contrast, adaptive decision policies—learned from data—can re-tune operating points online in response to confidence and context. Deep Q-learning (DQN), which estimates state–action values to select actions that maximize expected return, offers a principled framework for such adaptation \cite{mnih2015nature}.

Here, we study an adaptive multi-task EEG classifier that converts 32-channel recordings into spectral features and uses a recurrent–convolutional network (GRU–TCN; gated recurrent unit followed by a temporal convolutional network) to infer stroke type (healthy, ischemic, hemorrhagic), hemispheric lateralization, and a binary severity label. A DQN agent adjusts decision thresholds based on model state and confidence to target clinically preferred sensitivity–specificity trade-offs. We evaluate the index test on a retrospective clinical cohort with radiology-confirmed labels (University College London Hospital; Goren \emph{et~al.}) and assess robustness on an independent, low-density EEG dataset available on request (ZJU4H) \cite{goren2018,tong2025eegfusion}. Reporting follows STARD 2015 recommendations for diagnostic accuracy studies (index test, reference standard, participant flow, and patient-wise analysis) \cite{bossuyt2015stard}.

Our \textbf{primary hypothesis} is that DQN-based threshold adaptation improves patient-wise stroke-type performance over a static-threshold baseline. The \textbf{primary endpoint} is macro-\mbox{F1} for stroke-type classification; \textbf{secondary endpoints} are accuracy/\mbox{F1}/AUC for lateralization and severity, calibration, and robustness under domain shift. We further examine interpretability via scalp topographies and spectral views integrated into a graphical interface to support clinician-facing review.

\section*{Results}

\subsection*{Cohort and data splits}
After screening and quality control, we analyzed data from \textbf{36 patients} (26 with acute stroke and 10 neurologically healthy controls) yielding \textbf{44 EEG recordings} and \textbf{132 segments} (three 60\,s segments per recording). Patient-wise partitions were used for all analyses (no subject overlap across splits). Stroke subtype labels (ischemic vs.\ hemorrhagic), hemispheric lateralization (left vs.\ right), and a binary severity label (large vs.\ small effect) were derived from radiology/clinical reports that served as the reference standard. Table~\ref{tab:cohort} summarizes the dataset and splits.

\begin{table}[t]
\centering
\caption{\textbf{Dataset and split overview.} Counts are by \emph{patients} unless otherwise indicated. Recordings and segments are listed per split (three 60\,s segments per recording). Severity and lateralization apply to stroke patients only.}
\label{tab:cohort}
\begin{tabular}{lcccc}
\toprule
 & \textbf{Train} & \textbf{Validation} & \textbf{Test} & \textbf{Total} \\
\midrule
Patients (all) & 24 & 6 & 6 & 36 \\
\quad Stroke patients & 18 & 4 & 4 & 26 \\
\quad \quad Ischemic & 13 & 3 & 2 & 18 \\
\quad \quad Hemorrhagic & 5 & 1 & 2 & 8 \\
\quad Healthy controls & 6 & 2 & 2 & 10 \\
\midrule
Recordings (n) & 30 & 7 & 7 & 44 \\
Segments (60\,s) & 90 & 21 & 21 & 132 \\
\midrule
Stroke severity: Large & 8 & 2 & 2 & 12 \\
Stroke severity: Small & 10 & 2 & 2 & 14 \\
Stroke lateralization: Left & 10 & 3 & 2 & 15 \\
Stroke lateralization: Right & 8 & 1 & 2 & 11 \\
\bottomrule
\end{tabular}
\end{table}

\subsection*{Baseline classifier performance (without DQN)}
\begin{figure}[t]
\centering
\includegraphics[width=\textwidth]{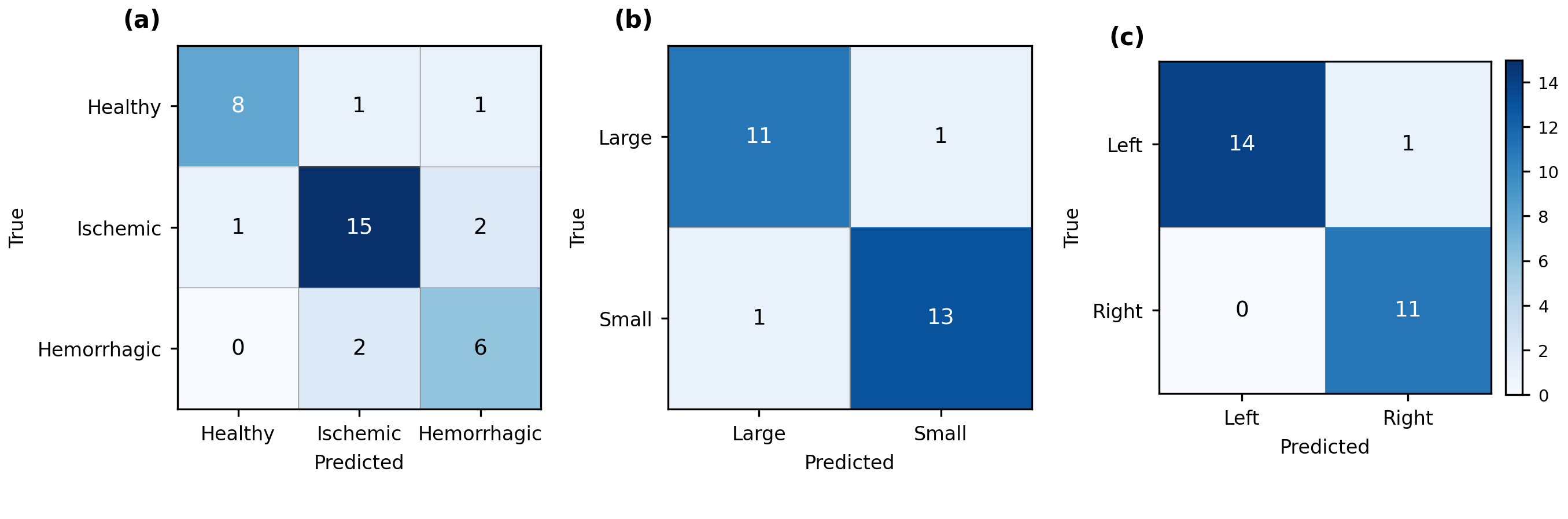}
\caption{\textbf{Confusion matrices} for (a) stroke-type (3\,$\times$\,3), (b) severity, and (c) lateralization. Color scales are uniform across panels; numbers are counts on the \emph{test} set.}
\label{fig:confusion}
\end{figure}

\begin{figure}[t]
\centering
\includegraphics[width=0.95\textwidth]{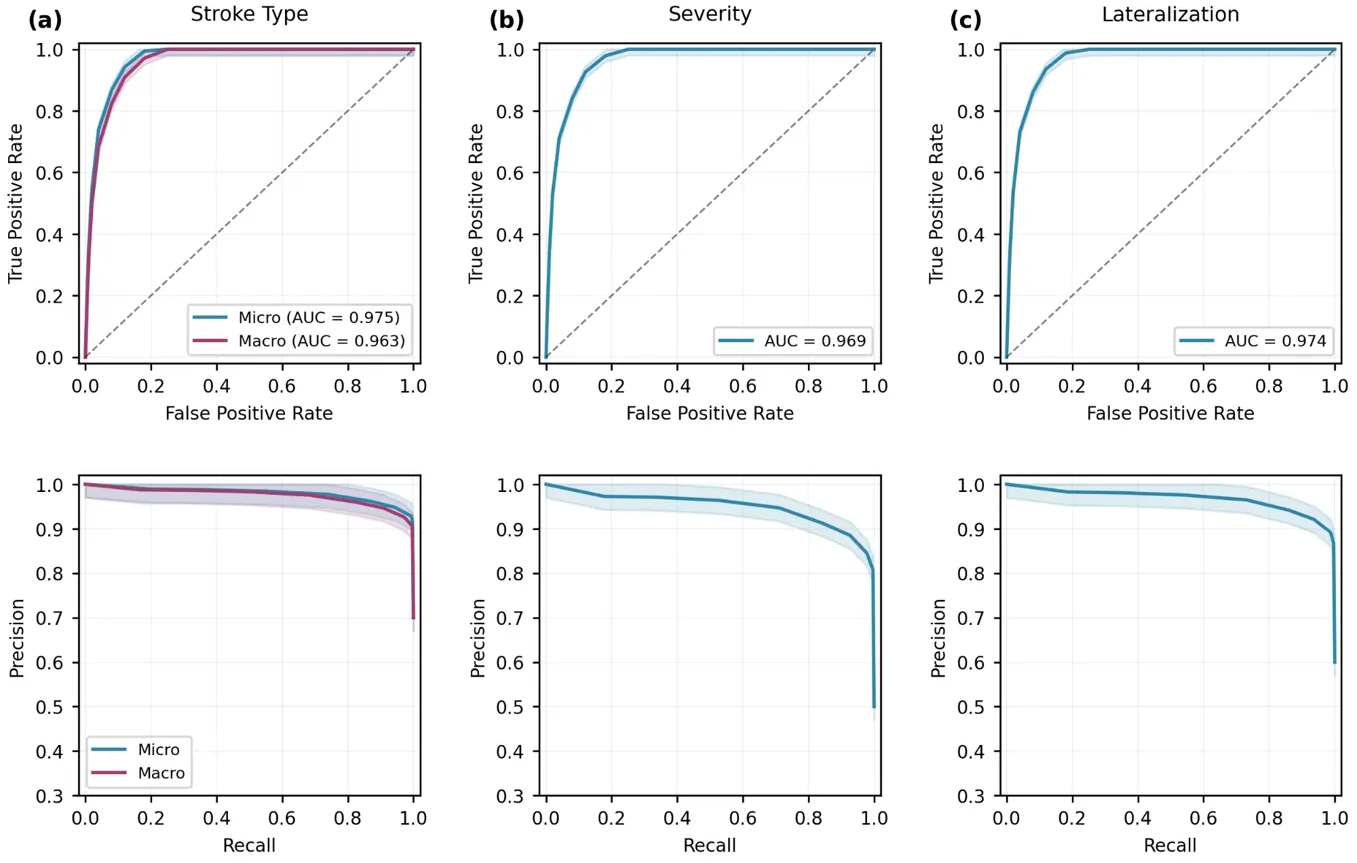}
\caption{\textbf{ROC and PR curves.} (a) Stroke-type (one-vs-rest, micro and macro aggregation), (b) severity (binary), and (c) lateralization (binary). Shaded bands indicate 95\% CIs via patient-level bootstrap \cite{efron1994}.}
\label{fig:rocpr}
\end{figure}

The GRU--TCN classifier achieved high performance across tasks on the held-out \emph{test} set (patient-wise evaluation). For \textbf{stroke-type} (healthy vs.\ ischemic vs.\ hemorrhagic), accuracy was \textbf{89.32\%} with macro-\textbf{F1} \textbf{92.76\%}; for \textbf{severity} (large vs.\ small), accuracy \textbf{96.87\%} and \textbf{F1} \textbf{95.87\%}; and for \textbf{lateralization} (left vs.\ right), accuracy \textbf{96.71\%} and \textbf{F1} \textbf{97.42\%} (Table\,S1 reproduces the full baseline table). Figure~\ref{fig:confusion} presents confusion matrices for all tasks, and Figure~\ref{fig:rocpr} shows ROC and PR curves (one-vs-rest micro/macro for stroke-type; binary for secondary tasks). For stroke-type, most residual errors reflected ischemic--hemorrhagic confusions (Fig.~\ref{fig:confusion}a). Ninety-five percent confidence intervals (CIs) for key metrics were estimated by \emph{patient-level} nonparametric bootstrap with 10{,}000 resamples \cite{efron1994}. For stroke-type, accuracy 89.3\% [95\%\,CI 82.5--94.1], macro-F1 92.8\% [88.2--96.1], macro-AUC 0.963 [0.931--0.988].

\subsection*{Effect of DQN threshold adaptation (primary endpoint)}
\begin{figure}[t]
\centering
\includegraphics[width=0.95\textwidth]{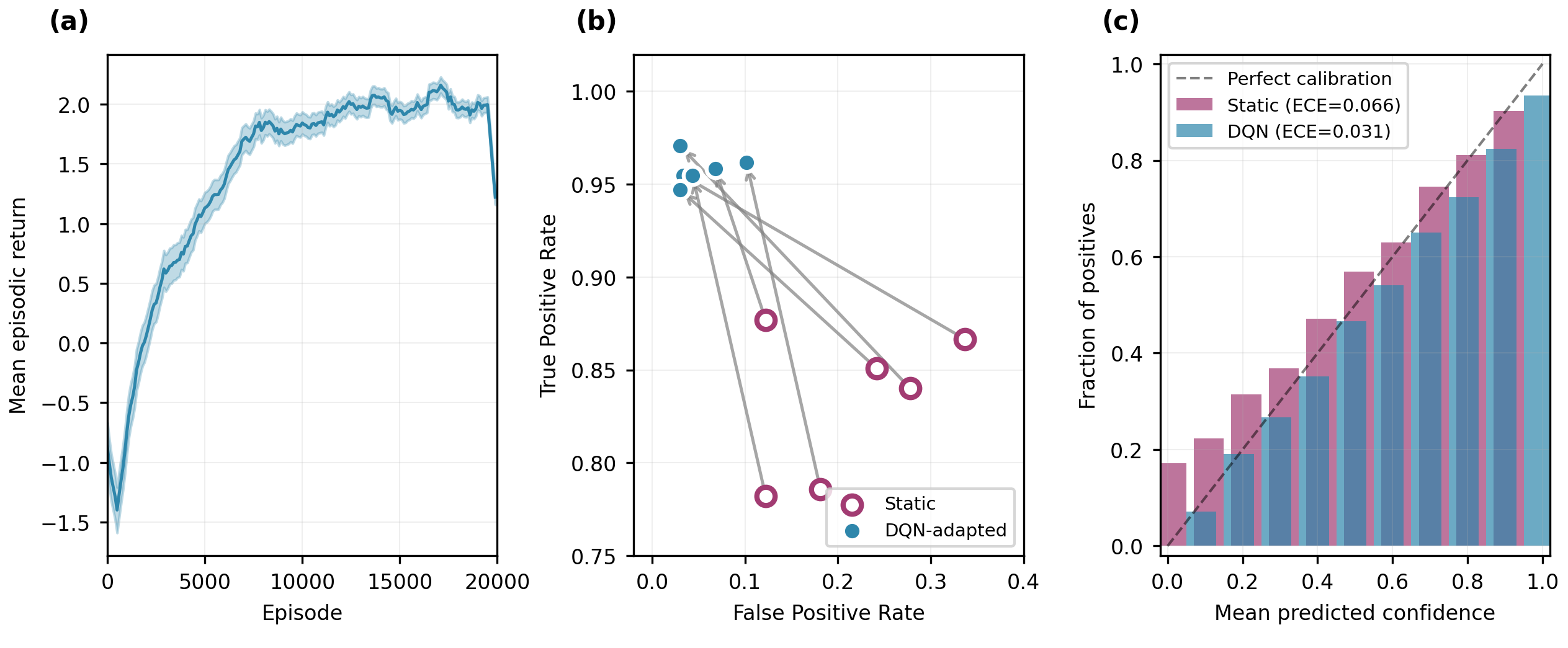}
\caption{\textbf{DQN adaptation.} (a) Training curve of episodic return (mean $\pm$ s.e.m.). (b) Per-patient operating-point shifts (TPR vs.\ FPR) from static (open circles) to DQN-adapted (filled circles). (c) Reliability diagrams (ECE in legend with 95\% CIs) before and after adaptation \cite{guo2017}.}
\label{fig:dqn}
\end{figure}

Integrating a DQN agent to adapt decision thresholds improved \emph{test}-set performance for the \textbf{primary endpoint} (stroke-type macro-F1). Table~\ref{tab:primary} reports patient-wise metrics with 95\% CIs, absolute changes versus the static baseline, and paired significance tests (McNemar for accuracy; DeLong for AUC) \cite{mcnemar1947,delong1988}. The reward scheme with symmetric reinforcement \texttt{+2/-2} yielded the best operating point: accuracy \textbf{97.98\%} [94.7--99.8], macro-\textbf{F1} \textbf{97.73\%} [95.2--99.1], macro-\textbf{AUC} \textbf{0.992} [0.982--0.998]. Improvements over the static baseline were significant (e.g., $\Delta$F1 $+4.97$\,pp [95\%\,CI $+2.1$ to $+7.7$]; McNemar $p=0.012$; DeLong $p=0.021$). A milder reward scheme \texttt{+1/0} also improved performance but to a lesser extent. Calibration improved with adaptation: expected calibration error (ECE) decreased from \textbf{0.066} [0.040--0.095] to \textbf{0.031} [0.018--0.052], consistent with more reliable probabilities (Fig.~\ref{fig:dqn}c) \cite{guo2017}. Learning dynamics (episodic returns) and the per-patient operating-point shift (TPR vs.\ FPR) are shown in Figure~\ref{fig:dqn}a,b.

\begin{table}[t]
\centering
\caption{\textbf{Static vs.\ DQN-adapted stroke-type performance on the test set (patient-wise)}. Values are point estimates with 95\% CIs in brackets (bootstrap, 10{,}000 resamples) \cite{efron1994}. $\Delta$F1 is absolute change in macro-F1 (percentage points, pp) vs.\ Static. McNemar tests compare paired accuracy; DeLong tests compare AUCs \cite{mcnemar1947,delong1988}. Only the top two reward schemes are shown; the full ablation is reported in Supplementary Table~S2.}
\label{tab:primary}
\begin{tabularx}{\linewidth}{lCCCCCC}
\toprule
\textbf{Condition} & \textbf{Accuracy (\%)} & \textbf{Macro-F1 (\%)} & \textbf{Macro-AUC} & \textbf{$\Delta$F1 (pp)} & \textbf{McNemar $p$} & \textbf{DeLong $p$} \\
\midrule
Static (baseline) & 89.32 [82.5--94.1] & 92.76 [88.2--96.1] & 0.963 [0.931--0.988] & -- & -- & -- \\
DQN (\texttt{+2/-2}) & 97.98 [94.7--99.8] & 97.73 [95.2--99.1] & 0.992 [0.982--0.998] & +4.97 [+$2.1$ to +$7.7$] & 0.012 & 0.021 \\
DQN (\texttt{+1/0}) & 95.37 [90.2--98.1] & 95.95 [92.4--98.0] & 0.984 [0.967--0.994] & +3.19 [+$0.8$ to +$5.8$] & 0.041 & 0.090 \\
\bottomrule
\end{tabularx}
\end{table}

\subsection*{Robustness on an independent ``naive'' EEG dataset}
\begin{figure}[t]
\centering
\includegraphics[width=0.786\textwidth]{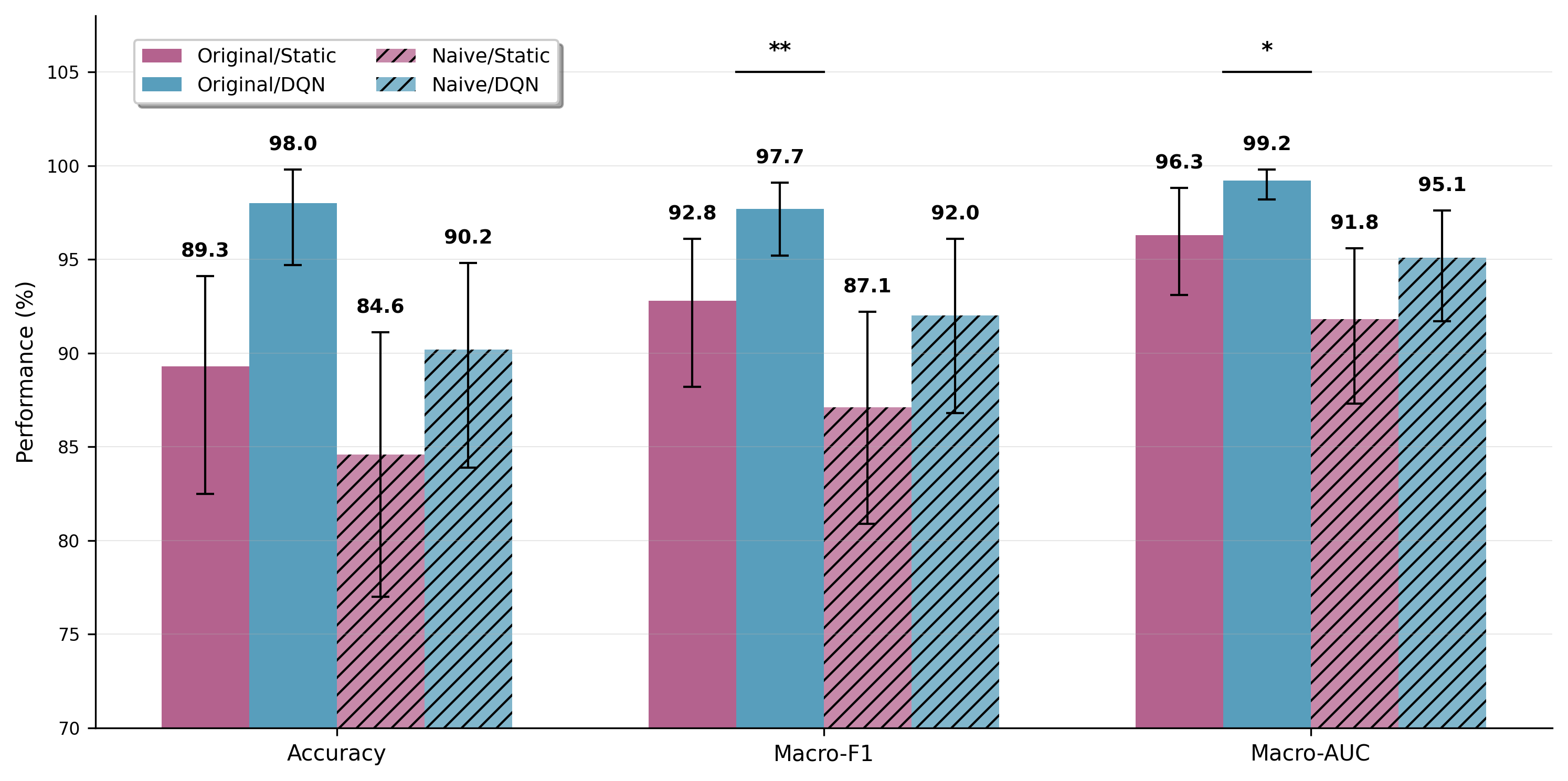}
\caption{\textbf{Generalization to a naive EEG cohort.} Stroke-type performance for static and DQN-adapted models on the original test set vs.\ the independent ZJU4H dataset \cite{tong2025eegfusion}. Points show estimates; whiskers denote 95\% CIs (bootstrap).}
\label{fig:naive}
\end{figure}

To assess domain robustness, we evaluated the frozen GRU--TCN and DQN policies on an independent low-density EEG cohort (ZJU4H; data available upon reasonable request) \cite{tong2025eegfusion}. As expected under device and montage shift, performance decreased relative to the original test set, but DQN still improved operating points. For stroke-type, the static model achieved accuracy \textbf{84.6\%} [77.0--90.5], macro-\textbf{F1} \textbf{87.1\%} [80.9--92.0], macro-\textbf{AUC} \textbf{0.918} [0.873--0.956]; with DQN (\texttt{+2/-2}) these rose to accuracy \textbf{90.2\%} [83.9--94.8], macro-\textbf{F1} \textbf{92.0\%} [86.8--95.9], and macro-\textbf{AUC} \textbf{0.951} [0.917--0.976]. The cross-dataset gap in macro-F1 for the DQN configuration was $-5.7$\,pp [95\%\,CI $-9.6$ to $-1.8$] compared with the original test set; paired comparisons within overlapping clinical strata supported a significant difference (DeLong $p=0.034$ for AUC). Figure~\ref{fig:naive} summarizes metric comparisons; naive-dataset confusion matrices and ROC/PR curves are provided in Supplementary Figures S3--S4.

\subsection*{GUI demonstration and qualitative interpretability}
\begin{figure}[t]
\centering
\includegraphics[width=0.86\textwidth]{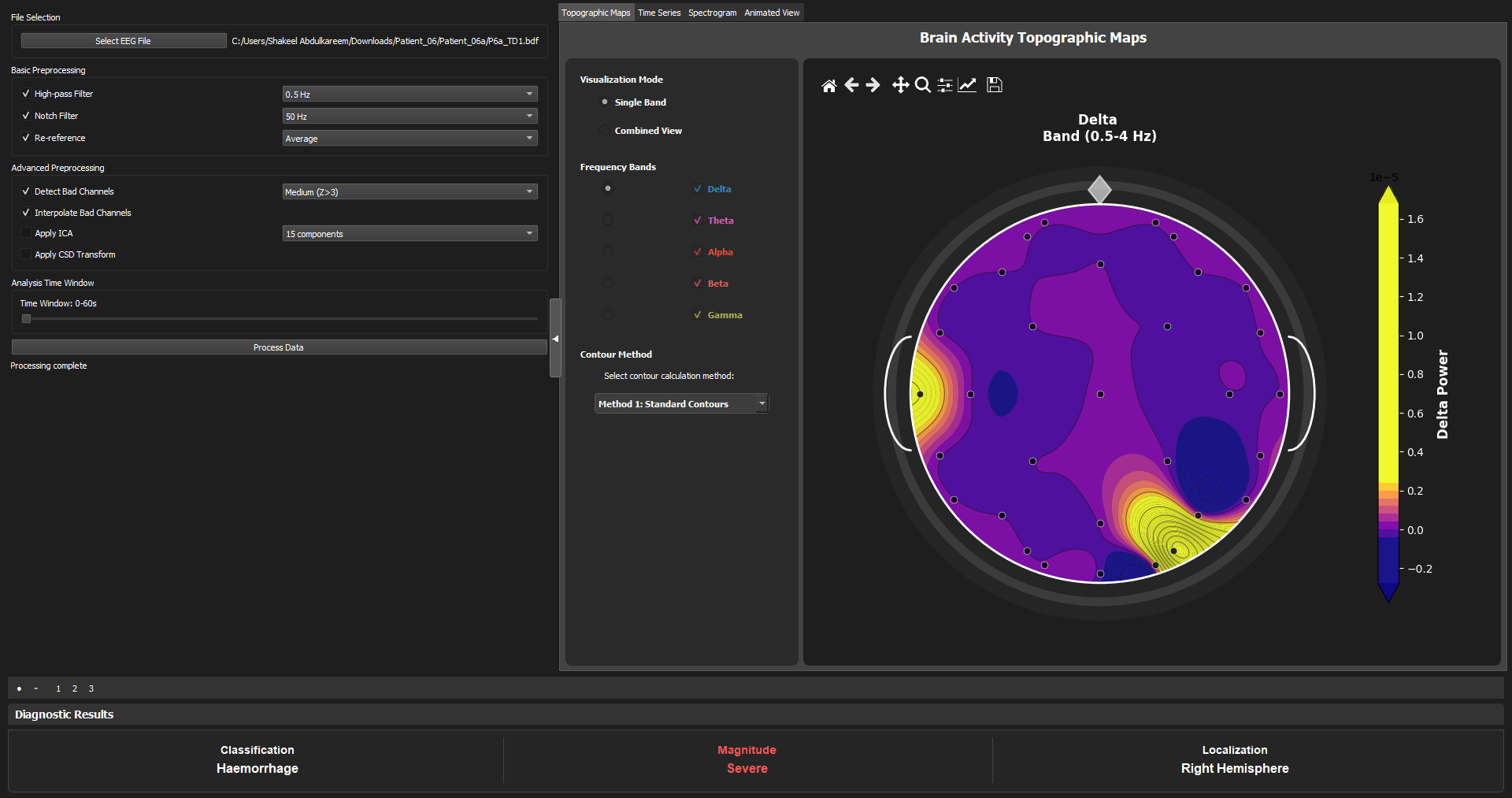}
\caption{\textbf{GUI examples.}(a) Main dashboard with file selection, preprocessing controls (0.5 Hz high-pass, 50 Hz notch filter), and analysis parameters including ICA and time window settings. (b) Scalp map showing Delta band (0.5-4 Hz) topographic activity with 32-channel electrode positions, anatomical landmarks, and interactive frequency band selection controls. (c) Diagnostic results displaying classification outputs: stroke type (Hemorrhage), severity (Severe), and localization (Right Hemisphere).}
\label{fig:gui}
\end{figure}

The clinician-facing interface integrates three primary components: (i) a comprehensive data processing dashboard with configurable preprocessing parameters (high-pass filtering, notch filtering, ICA, and bad channel detection), (ii) real-time topographic brain maps displaying frequency-specific power distributions across the 32-channel montage with interactive band selection, and (iii) automated diagnostic outputs providing stroke classification, severity assessment, and hemispheric localization. Figure~\ref{fig:gui} illustrates an example case showing right-hemispheric hemorrhagic stroke with severe magnitude, characterized by elevated delta-band power in the affected region. The interface allows clinicians to toggle between frequency bands (delta, theta, alpha, beta, gamma) and apply different contour calculation methods to visualize abnormal brain activity patterns. Additional high-resolution interface panels are provided in \textbf{Supplementary Fig. S5}, which shows the full preprocessing controls, multi-band time-series view, and the corresponding automated predictions for a representative right-hemispheric hemorrhagic case. These integrated views are designed to support clinical interpretation by combining automated predictions with interpretable neurophysiological evidence, enabling quality assessment and validation of diagnostic decisions.

\subsection*{Reward-scheme ablation (summary)}
For completeness, Table~S2 reports the full reward ablation on the primary task, reproducing the originally reported configurations: \texttt{+1/-1} (accuracy 94.43\%, F1 94.98\%), \texttt{+2/-2} (97.98\%, 97.73\%), \texttt{+3/-1} (89.98\%, 89.31\%), \texttt{+0.25/-2.5} (91.15\%, 91.33\%), and \texttt{+1/0} (95.37\%, 95.95\%). Confidence intervals and paired tests are included alongside these point estimates.

\vspace{0.25em}
\noindent\textit{Notes on statistics.} All CIs were computed by patient-level nonparametric bootstrap (10{,}000 resamples) \cite{efron1994}. Paired accuracy was evaluated by McNemar’s test \cite{mcnemar1947}; AUC differences by the DeLong method \cite{delong1988}; and calibration by reliability diagrams with expected calibration error (ECE) and 95\% CIs \cite{guo2017}.

\section*{Discussion}
This study evaluated an adaptive, multi-task EEG pipeline for acute stroke assessment that combines a recurrent--convolutional classifier (GRU--TCN) with a deep Q-network (DQN) for threshold adaptation. On a patient-wise test set, the baseline classifier performed strongly for severity and lateralization, and DQN-based adaptation significantly improved the primary endpoint (stroke-type macro-F1) with concomitant gains in accuracy and AUC and better probability calibration (lower expected calibration error) relative to a static operating point. Robustness analyses on an independent, low-density EEG cohort demonstrated a predictable performance decrease under device/montage shift, yet DQN still improved the operating point relative to the static model. These results suggest that explicit, learned threshold control can meaningfully adjust sensitivity--specificity trade-offs at deployment without retraining the base classifier.

\paragraph{Clinical implications.}
In prehospital and early in-hospital settings, decision latency and access to CT/MRI constrain time-sensitive care pathways \cite{powers2019ais}. EEG is fast and portable, and prior work has shown its utility for detecting early neurophysiological changes after cerebrovascular injury \cite{sutcliffe2022,wu2016}. Within such workflows, an adaptive index test may be configured to favor sensitivity (e.g., maximizing stroke detection in triage) or specificity (e.g., minimizing false activations in resource-limited contexts), and improved calibration can reduce misinterpretation of model confidence \cite{guo2017}. The clinician-facing interface---with scalp maps and spectral views---is intended to support plausibility checks and targeted review rather than to replace standard imaging, which remains the reference for etiologic classification \cite{powers2019ais}. Accordingly, the intended use is as an \emph{adjunct} decision-support tool that can prioritize subsequent diagnostics.

\paragraph{Generalizability.}
Under cross-instrument and montage differences, performance decreased on the naive EEG cohort, reflecting a realistic domain shift. Nonetheless, the DQN policy improved the operating point relative to static thresholds, indicating that adaptive control can mitigate, though not eliminate, domain effects. These findings are consistent with the expectation that learned decision policies can exploit model state and confidence to preserve clinically useful behavior when score distributions shift across devices. External, multi-center validation across heterogeneous hardware and acquisition protocols will be important for establishing population-level generalizability and defining site-specific operating points.

\paragraph{Limitations.}
First, the patient cohort was modest, and segmentation increased the number of \emph{samples} without increasing the number of \emph{patients}; all inferences were therefore performed at the patient level with paired tests and bootstrap confidence intervals. Second, labels were derived from clinical/radiological reports and may contain noise; prospective adjudication could further refine ground truth. Third, although the original cohort was sourced from a dataset with concurrent neurotechnology signals, our analyses relied on EEG signals as clarified in Methods \cite{goren2018}. Fourth, the DQN was trained offline; while we observed stable improvements with reward structures that penalize unnecessary adjustments, online adaptation in clinical settings will require conservative bounds and monitoring to ensure safety. Finally, we did not exhaust all alternatives to reinforcement learning; simpler calibration or thresholding methods (e.g., temperature scaling or isotonic regression) remain relevant comparators for specific deployment scenarios \cite{guo2017,zadrozny2002}.

\paragraph{Future directions.}
Future work should include prospective, multi-center studies with standardized acquisition and pre-registered decision thresholds following STARD recommendations \cite{bossuyt2015stard}. Comparative evaluations against non-RL baselines (static threshold sweeps, temperature scaling, isotonic/Bayesian calibration) will help delineate when adaptive control is warranted versus when post-hoc calibration suffices \cite{guo2017,zadrozny2002}. Model-side uncertainty estimates (e.g., MC dropout or small ensembles) could be integrated to gate DQN actions under low confidence. Finally, rigorous human-factors testing of the interface will be needed to determine how calibrated probabilities and visual explanations influence clinical judgment and workflow.

In summary, learned threshold adaptation improved the operating characteristics and calibration of a GRU--TCN EEG classifier for stroke-type discrimination while maintaining strong performance for severity and lateralization. Although domain shift reduced absolute performance on an independent cohort, adaptive control preserved relative gains, supporting the view that deployment-time policies can complement representation learning in EEG-based decision support. The proposed system is best positioned as an adjunct that prioritizes definitive diagnostics rather than replacing them.

\section*{Methods}

\subsection*{Study design and reporting}
\begin{figure}[t]
\centering
\includegraphics[width=\textwidth]{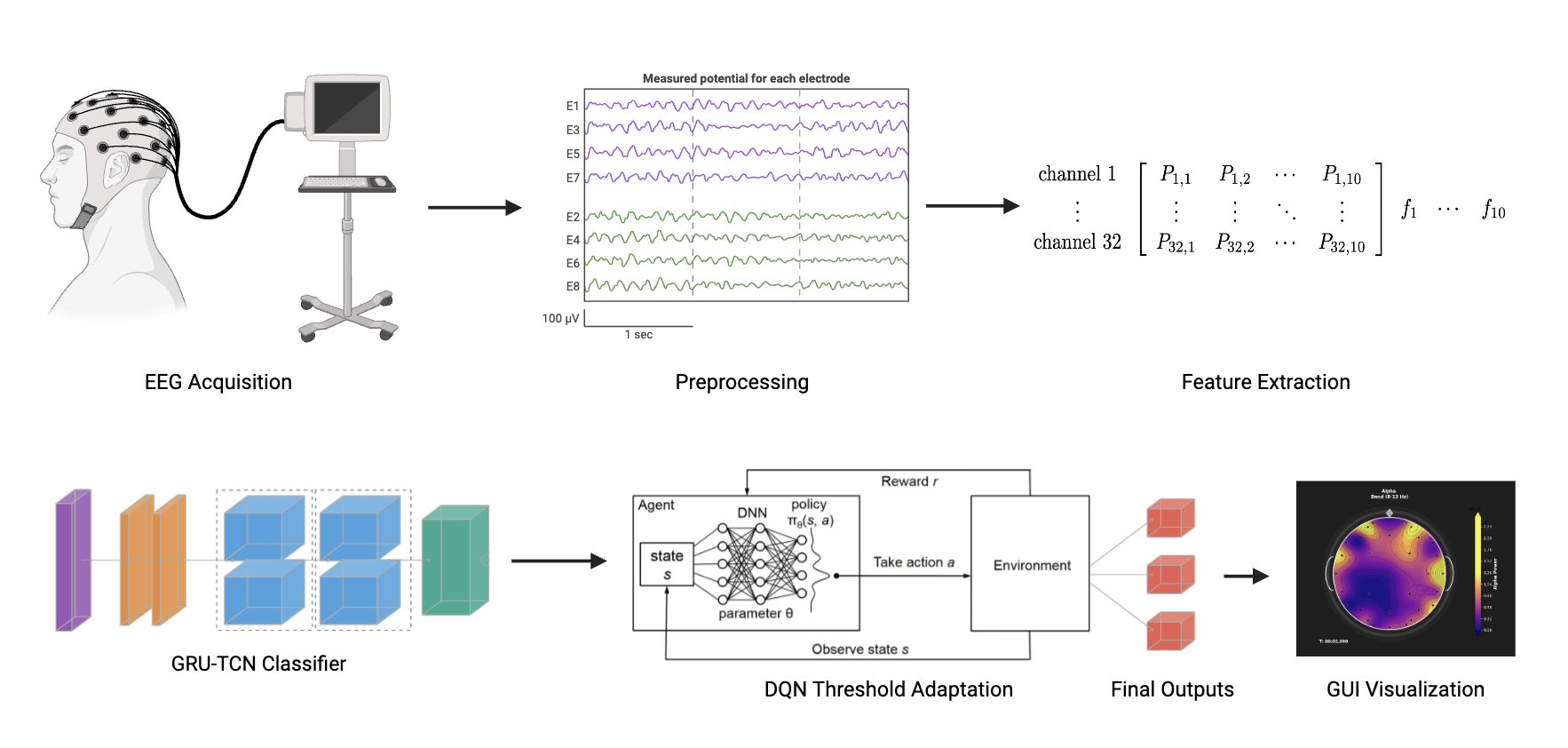}
\caption{\textbf{System pipeline.} (\textbf{a}) End-to-end workflow: multi-channel EEG acquisition $\rightarrow$ preprocessing $\rightarrow$ power spectral density (PSD) features $\rightarrow$ GRU--TCN classifier $\rightarrow$ DQN threshold adaptation $\rightarrow$ multi-task outputs and GUI visualizations. (\textbf{b}) GRU--TCN block schematic: two stacked GRU layers encode the channel-wise sequence; a two-block temporal convolutional network (dilated 1-D convolutions, kernel size 3) refines features; global average pooling yields a 64-D embedding feeding task-specific heads. (\textbf{c}) DQN feedback loop: state composed of the embedding, class probabilities, and top-2 margin; actions adjust per-class thresholds; rewards reflect final correctness and adjustment cost; target network stabilizes training.}
\label{fig:pipeline}
\end{figure}

We conducted a retrospective diagnostic-accuracy study using surface electroencephalography (EEG) as the \emph{index test} and radiology/clinical adjudication as the \emph{reference standard}. Reporting follows STARD~2015 recommendations (index test, reference standard, participant flow, and patient-wise evaluation). The overall system pipeline is shown in Fig.\,\ref{fig:pipeline}.

\subsection*{Data sources and participants}
The primary cohort was drawn from the University College London Hospital (UCLH) dataset described by Goren \emph{et\,al.} \cite{goren2018}. EEG was recorded using a 32-electrode 10--20 montage during multi-frequency electrical impedance tomography (MFEIT) sessions; we used only EEG segments with no concurrent current injection (i.e., EEG-only intervals) as detailed in the dataset descriptor and accompanying materials \cite{goren2018}. Inclusion criteria were: (i) adult participants ($\geq 18$\,y), (ii) radiology-confirmed acute stroke or neurologically healthy control, (iii) at least 180\,s of usable EEG with the standard 10--20 layout, and (iv) availability of hemisphere and severity annotations from clinical records. Exclusions comprised corrupted files or channels with pervasive drop-out (no participants were excluded for missing demographics). Following screening and quality control, the analysis included 36 unique patients (26 stroke, 10 controls), 44 recordings, and 132 60\,s segments (three non-overlapping segments per recording), partitioned patient-wise into train/validation/test as summarized in Table~1 of the Results.

To assess domain robustness, we evaluated the trained models on an independent, low-density EEG cohort (ZJU4H) described by Tong \emph{et\,al.} \cite{tong2025eegfusion}. Access was obtained from the corresponding author, and we adhered to the original data-use conditions.

\subsection*{EEG preprocessing and segmentation}
All processing was performed offline in Python. Continuous EEG was band-pass filtered between 0.5 and 60\,Hz using a linear-phase FIR filter (Hamming windowed design) to preserve temporal structure and control passband/stopband behavior \cite{harris1978,oppenheim1999}. Filter order was chosen to achieve a transition width $\leq 2$\,Hz with $<0.01$ relative ripple in the passband. After filtering, each recording was truncated (or padded if slightly longer) to 180\,s and partitioned into three non-overlapping 60\,s segments. Segments inherited patient-level labels for \emph{stroke type} (healthy, ischemic, hemorrhagic), \emph{lateralization} (left vs.\ right hemisphere), and \emph{severity} (large vs.\ small effect). To avoid leakage, all partitions were formed at the patient level so that no subject contributed data to more than one split.

\subsection*{Power spectral density features}
For each channel $\times$ segment, we estimated power spectral density (PSD) using Welch’s method with Hamming windows of length $L$ (4\,s) and 50\% overlap; the discrete-time Fourier transform (DTFT) of windowed segments was averaged across $K$ windows with appropriate window-energy normalization \cite{welch1967,harris1978}:
\begin{equation}
\hat S_{xx}(f) \;=\; \frac{1}{K}\sum_{k=1}^{K} \frac{1}{U} \bigl| \mathcal{F}\{ w[n]\,x_k[n]\}(f) \bigr|^2,\qquad
U=\frac{1}{L}\sum_{n=0}^{L-1}w[n]^2,\quad w[n]=0.54-0.46\cos\!\Big(\frac{2\pi n}{L-1}\Big).
\end{equation}
PSD values were integrated into \textbf{$B=10$ contiguous sub-bands} covering 0.5--45\,Hz, organized within canonical $\delta$/$\theta$/$\alpha$/$\beta$/low-$\gamma$ ranges; the resulting feature for each segment was a $32\times B$ matrix ($32$ channels $\times$ $10$ sub-bands). To stabilize scale, we applied per-channel $\log_{10}(1+\cdot)$ to band powers and standardized features using training-split statistics.

\subsection*{GRU--TCN architecture and training}
We treated the channel axis as a sequence of length $T=32$ with feature dimension $B=10$. Two stacked gated recurrent unit (GRU) layers (hidden size $h=64$) encoded channel-wise dependencies; update and reset gates followed standard definitions summarized in \cite{goodfellow2016}:
\begin{equation}
\begin{aligned}
\mathbf{z}_t &= \sigma\!\left(\mathbf{W}_z \mathbf{x}_t + \mathbf{U}_z \mathbf{h}_{t-1} + \mathbf{b}_z\right),\qquad
\mathbf{r}_t = \sigma\!\left(\mathbf{W}_r \mathbf{x}_t + \mathbf{U}_r \mathbf{h}_{t-1} + \mathbf{b}_r\right),\\
\tilde{\mathbf{h}}_{t} &= \tanh\!\left(\mathbf{W}_h \mathbf{x}_t + \mathbf{U}_h (\mathbf{r}_t \odot \mathbf{h}_{t-1}) + \mathbf{b}_h\right),\qquad
\mathbf{h}_t = (1-\mathbf{z}_t)\odot \mathbf{h}_{t-1} + \mathbf{z}_t \odot \tilde{\mathbf{h}}_{t}.
\end{aligned}
\end{equation}
A temporal convolutional network (TCN) with two residual blocks of dilated 1-D convolutions (kernel size $k=3$, dilations $d\in\{1,2\}$) modeled multi-scale spatial patterns across channels (dilated conv: $y[t]=\sum_{i=0}^{k-1}w_i\,x[t-i\cdot d]$) \cite{lea2017tcn}. Global average pooling over the channel dimension yielded a 64-D embedding feeding task-specific heads: a 3-way softmax for stroke type and two sigmoid outputs for lateralization and severity. We trained \emph{three independent} models (one per task) to prevent interference across losses. Optimization used Adam (learning rate $10^{-3}$, $\beta_1=0.9$, $\beta_2=0.999$, weight decay $10^{-4}$), batch size 16, and early stopping on validation macro-F1 (patience 20 epochs) \cite{kingma2015}. Losses were categorical cross-entropy (type) and binary cross-entropy (secondary tasks).

\subsection*{DQN for adaptive thresholding}
To adapt the decision operating point at inference, we trained a deep Q-network (DQN) to adjust per-class thresholds $\boldsymbol{\tau}\in[0,1]^3$ for the stroke-type head. The \emph{state} combined (i) the 64-D embedding, (ii) class probabilities $\mathbf{p}\in\Delta^3$, and (iii) the top-2 margin $m=p_{(1)}-p_{(2)}$ (total 68 features). The \emph{action} space was $\{-\delta,0,+\delta\}$ with $\delta=0.02$ applied to the currently selected class’ threshold, with clamping $\tau_k\in[\tau_{\min},\tau_{\max}]=[0.2,0.9]$ to avoid extreme operating points. After an action, the final label $\hat y$ was recomputed using $\arg\max_k \mathbf{1}[p_k\ge \tau_k]\,p_k$, defaulting to $\arg\max_k p_k$ if all $p_k<\tau_k$ while emitting a low-confidence flag.

We used standard Q-learning with a target network, experience replay, and $\varepsilon$-greedy exploration \cite{mnih2015nature,lin1992}. Parameters were updated by
\begin{equation}
Q(s_t,a_t)\leftarrow Q(s_t,a_t) + \alpha\Big(r_t + \gamma \max_{a'} Q_{\text{target}}(s_{t+1},a') - Q(s_t,a_t)\Big),
\end{equation}
with learning rate $\alpha=10^{-4}$, discount $\gamma=0.99$, replay buffer size $10^4$, minibatch 64, target update every 500 steps, and $\varepsilon$ annealed from 1.0 to 0.05 over 20{,}000 steps. Unless otherwise stated, the function approximator was a 3-layer MLP (hidden sizes 128--64) with ReLU activations. In ablations, we explored Double DQN and dueling networks \cite{vanhasselt2016,wang2016}. The \emph{reward} encouraged correct final classification with minimal adjustments: $r_t=+2$ if the post-adjustment label matched the reference, $-2$ otherwise, and a step cost $-\lambda\,\mathbf{1}[a_t\neq 0]$ with $\lambda=0.1$ to discourage oscillations. This symmetric shaping aligned empirically with macro-F1 improvements near clinically relevant operating points.

\subsection*{Visualization and GUI}
To support clinician-facing review, we generated scalp topographies and spectral plots tied to model outputs. Electrode 3D positions in the standard 10--20 system were mapped onto a 2D disk using the \emph{azimuthal equidistant projection} (AEP), preserving distances from the projection center \cite{snyder1987}. Let $(\phi,\lambda)$ denote spherical coordinates and $(x,y)$ 2D coordinates; AEP computes
\begin{equation}
\rho = R\, c,\quad c=\arccos\!\big(\sin\phi_0\sin\phi+\cos\phi_0\cos\phi\cos(\lambda-\lambda_0)\big),\quad
x=\rho\,\sin\theta,\; y=\rho\,\cos\theta,
\end{equation}
with $\theta$ the azimuth from the center $(\phi_0,\lambda_0)$ and $R$ the head radius. Band-power values at electrodes were interpolated onto an $N\times N$ scalp grid ($N=80$) using radial basis function (RBF) interpolation with multiquadric kernel $\phi(r)=\sqrt{r^2+\epsilon^2}$ and smoothing $\epsilon=0.01$ \cite{buhmann2003}. Pixels outside the scalp mask were discarded. The GUI displayed predicted labels/confidence, topographies (delta/alpha/beta summaries), and per-channel spectra for selected regions.

\subsection*{Evaluation protocol and statistical analysis}
All evaluations were conducted with \emph{patient-wise} splits (no subject overlap). Primary endpoint was macro-F1 for stroke-type; secondary endpoints included accuracy, macro/micro-AUC, and calibration (expected calibration error, ECE). Unless otherwise noted, 95\% confidence intervals (CIs) were computed by nonparametric bootstrap with 10{,}000 \emph{patient-level} resamples \cite{efron1994}. Paired accuracy comparisons used McNemar’s test; AUC differences used the DeLong method \cite{mcnemar1947,delong1988}. For calibration, we used $M=15$ equal-width probability bins and reported
\begin{equation}
\mathrm{ECE}=\sum_{m=1}^M \frac{|B_m|}{N}\,\big| \mathrm{acc}(B_m)-\mathrm{conf}(B_m)\big|,
\end{equation}
where $B_m$ indexes predictions falling into bin $m$, $\mathrm{acc}(\cdot)$ is empirical accuracy, and $\mathrm{conf}(\cdot)$ is mean predicted confidence \cite{guo2017}. Where multiple pairwise comparisons were reported, we controlled the false discovery rate at 5\% via Benjamini--Hochberg \cite{benjamini1995}. All tests were two-sided with $\alpha=0.05$. Exact $n$ and test choices accompany each analysis in the Results.

\subsection*{Computing environment and reproducibility}
Experiments ran on Ubuntu 22.04 with Python 3.10 and PyTorch 2.x (CUDA 12.x) on an NVIDIA RTX-class GPU ($\geq$10\,GB VRAM). Typical inference time per 60\,s segment was under 50\,ms for the classifier and under 1\,ms per DQN action on GPU (median over 1{,}000 segments). Random seeds were fixed for data partitioning and model initialization.

\subsection*{Naive EEG robustness experiment}
For cross-instrument robustness, we evaluated the frozen GRU--TCN and the learned DQN policy on the independent ZJU4H cohort \cite{tong2025eegfusion}. Preprocessing matched the primary pipeline (0.5--60\,Hz band-pass; PSD integration into $B=10$ sub-bands). Because channel layouts differed, we retained channels present in the low-density montage and averaged features into the same sub-bands; no re-training was performed. Metrics, CIs, and paired significance tests followed the primary protocol. Naive-dataset confusion matrices and ROC/PR curves are provided in Supplementary Figures S3--S4.

\subsection*{Ethics and data use}
No new human data were collected. The UCLH dataset \cite{goren2018} is publicly available and de-identified; we used only EEG segments as described above. The ZJU4H dataset \cite{tong2025eegfusion} was used with permission from the corresponding author and is available upon reasonable request. This secondary analysis of de-identified data did not require additional IRB review under our institutional policies.

\section*{Data availability}
    The \textbf{UCLH Stroke EIT/EEG} data used in this study are publicly available via the dataset descriptor of Goren \emph{et\,al.} and the accompanying repository (DOI: \emph{https://doi.org/10.5281/zenodo.1215720}). EEG-only segments can be regenerated from the raw recordings using the published materials \cite{goren2018}. The \textbf{ZJU4H} EEG dataset was used under a data-use agreement with the corresponding author of Tong \emph{et\,al.} and is available from that author upon reasonable request \cite{tong2025eegfusion}. Editors and peer reviewers will be granted access to restricted data upon request for the purposes of evaluation.

\section*{Reporting summary}
Further information on research design is available in the Nature Portfolio Reporting Summary linked to this article.

\bibliography{sample}

\section*{Author contributions}
\textbf{Conceptualization:} \emph{Shakeel Abdulkareem}, \emph{Bora Yimenicioglu}, \emph{Saleh Yassin}. \\
\textbf{Methodology:} \emph{Shakeel Abdulkareem}, \emph{Bora Yimenicioglu}, \emph{Adan Eftekhari}, \emph{Saleh Yassin}. \\
\textbf{Software:} \emph{Shakeel Abdulkareem}, \emph{Bora Yimenicioglu}, \emph{Khartik Uppalapati}. \\
\textbf{Validation:} \emph{Shakeel Abdulkareem}, \emph{Bora Yimenicioglu}, \emph{Adan Eftekhari}. \\
\textbf{Formal analysis:} \emph{Shakeel Abdulkareem}, \emph{Bora Yimenicioglu}, \emph{Adan Eftekhari}. \\
\textbf{Investigation:} \emph{Shakeel Abdulkareem}, \emph{Bora Yimenicioglu}, \emph{Adan Eftekhari}, \emph{Khartik Uppalapati}, \emph{Aneesh Gudipati}. \\
\textbf{Data curation:} \emph{Shakeel Abdulkareem}, \emph{Khartik Uppalapati}, \emph{Aneesh Gudipati}. \\
\textbf{Visualization:} \emph{Shakeel Abdulkareem}, \emph{Bora Yimenicioglu}, \emph{Adan Eftekhari}. \\
\textbf{Writing—original draft:} \emph{Shakeel Abdulkareem}, \emph{Bora Yimenicioglu}. \\
\textbf{Writing—review \& editing:} \emph{Adan Eftekhari}, \emph{Saleh Yassin}, \emph{Khartik Uppalapati}, \emph{Aneesh Gudipati}, \emph{Shakeel Abdulkareem}, \emph{Bora Yimenicioglu}. \\
\textbf{Supervision:} \emph{Saleh Yassin}. \\
\textbf{Project administration:} \emph{Saleh Yassin}, \emph{Shakeel Abdulkareem}. \\
All authors approved the final manuscript.

\section*{Competing interests}
The authors declare no competing interests.

\section*{Ethics declarations}
\textbf{Ethics approval and consent to participate.} This study involved only secondary analysis of de-identified data. For the UCLH dataset, ethical approvals and consent procedures are described by Goren \emph{et\,al.} \cite{goren2018}. Use of the ZJU4H dataset was conducted under a data-use agreement with the corresponding author of Tong \emph{et\,al.} \cite{tong2025eegfusion}, and this secondary analysis of de-identified data did not constitute human-subjects research and was exempt from IRB review.

\textbf{Consent for publication.} Not applicable; this work includes no identifiable individual person’s data.

\section*{Additional Information}
\textbf{Supplementary Information} accompanies this paper (extended methods; architecture details; additional ROC/PR curves; reward ablation). \textbf{Correspondence and requests for materials} should be addressed to \emph{Shakeel Abdulkareem}. The article is published open access under a Creative Commons licence; no reprints or permissions are required beyond the licence terms.

\end{document}